%% file: acl2023.tex
\pdfoutput=1

\documentclass[11pt]{article}

\usepackage[]{ACL2023}

\usepackage{times}
\usepackage{latexsym}
\usepackage{graphicx}
\usepackage{enumitem}
\usepackage[T1]{fontenc}

\usepackage[utf8]{inputenc}
\input{macro}

\usepackage{microtype}

\usepackage{inconsolata}
\usepackage{adjustbox}

\title{Translation or Recitation? Calibrating Evaluation Scores for Machine Translation of Extremely Low-Resource Languages}

\author{Danlu Chen$^{1}$, Ka Sing He$^{1}$, Jiahe Tian$^{2}$, Chenghao Xiao$^{3}$, Zhaofeng Wu$^{4}$,\\ \textbf{Taylor Berg-Kirkpatrick}$^{1}$, \textbf{Freda Shi}$^{5,6}$ \\
   UC San Diego$^{1}$,  New York University$^{2}$, Durham University$^{3}$,  \\
   MIT$^{4}$, University of Waterloo$^{5}$, Vector Institute$^{6}$ \\
  \texttt{dac013@ucsd.edu} 
}

\begin{document}
\maketitle
\begin{abstract}
The landscape of extremely low-resource machine translation (MT) is characterized by perplexing variability in reported performance, often making results across different language pairs difficult to contextualize. 
For researchers focused on specific language groups---such as ancient languages---it is nearly impossible to determine if breakthroughs reported in other contexts (e.g., native African or American languages) result from superior methodologies or are merely artifacts of benchmark collection. 
To address this problem, we introduce the \textbf{FRED Difficulty Metrics}, which include the \textit{Fertility Ratio (F)}, \textit{Retrieval Proxy} ($R$), \textit{Pre-training Exposure} ($E$), and \textit{Corpus Diversity} ($D$) and serve as dataset-intrinsic metrics to contextualize reported scores. 
These metrics reveal that a significant portion of result variability is explained by train-test overlap and pre-training exposure rather than model capability. 
Additionally, we identify that some languages — particularly extinct and non-Latin indigenous languages — suffer from poor tokenization coverage (high token fertility), highlighting a fundamental limitation of transferring models from high-resource languages that lack a shared vocabulary. 
By providing these indices alongside performance scores, we enable more transparent evaluation of cross-lingual transfer and provide a more reliable foundation for the XLR MT community.\footnote{\url{https://github.com/taineleau/FRED-loresMT/}}

\end{abstract}

\begin{figure}[t]
\centering
    \includegraphics[width=1\linewidth]{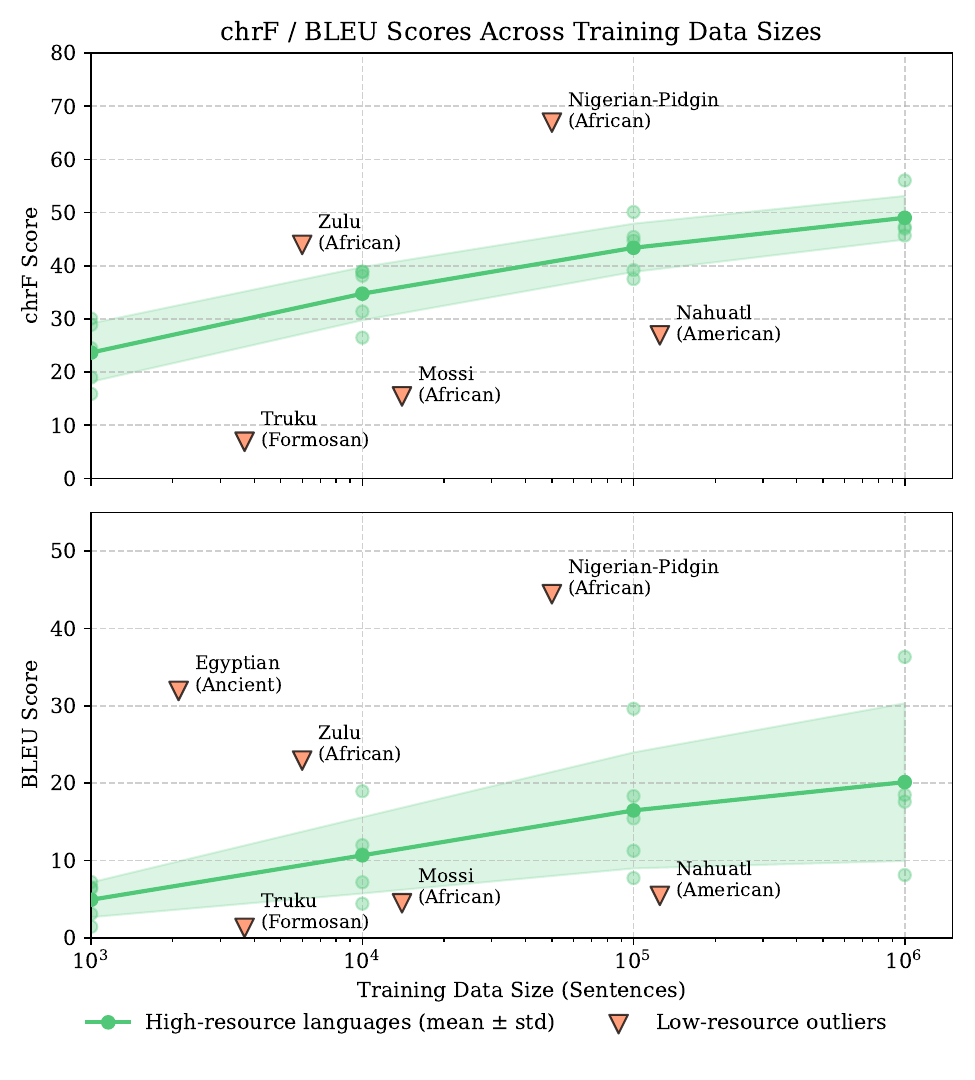}
    \vspace{-1em}
    \caption{Performance distribution of extremely low-resource (XLR) languages in machine translation, shown as scattered outliers. 
        The green line represents the mean performance of mBART when fine-tuned on varying sizes of training data from five high-resource languages (see Table \ref{tb:high_res} and \cref{ch:high_res} for details), with the shaded region indicating $\pm1$ standard deviation.     \label{fig:overview}
    }
\end{figure}

\section{Introduction}

Multilingual pre-trained models have significantly advanced machine translation for low-resource languages through cross-lingual transfer learning \cite{nllb2024scaling}. However, performance across extremely low-resource settings exhibits a staggering degree of variation. Recent studies \cite{haddow-etal-2022-survey} show that while some African and ancient languages can achieve BLEU scores exceeding 40, certain American or South Asian indigenous languages struggle to reach 5 BLEU under similar settings, i.e., translating to a high-resource language with a similar amount of training data. 

This disparity creates a significant barrier for the MT community \cite{silva-etal-2024-benchmarking}. 
Without a standardized way to provide context for these results, it is difficult to interpret whether a high BLEU score indicates an effective model or a benchmark with unnaturally low complexity. 
The issue goes beyond metric choices: while BLEU is not suitable enough for morphologically rich languages, we identify a similar trend with ChrF. 
These observations raise a fundamental question: \textit{Is the variability in MT performance due to inherent linguistic properties, or is it an artifact of how benchmark datasets were collected?}

To answer this question, we consider a specific category: \textbf{Extremely Low-Resource (XLR)} languages. We characterize XLR languages as those that exist in the ``blind spot'' of current multilingual models---lacking either the monolingual data required for effective pre-training or the lexical (subword) overlap necessary for efficient transfer learning \cite{haddow-etal-2022-survey}. To establish a performance baseline, we cap high-resource language datasets to extremely low-resource sizes (ranging from $10^3$ to $10^6$ sentences). As illustrated in Figure \ref{fig:overview}, we plot the performance of mBART \cite{mbart} on five typologically diverse high-resource languages as a reference region. 

We find that the performances of many XLR languages act as significant outliers: some vastly overperform relative to their size, while others achieve scores much lower than the baseline. 
We hypothesize that train-test data relationships, not just training size or architecture, drive these discrepancies. 
To verify the hypothesis, we propose \textbf{Difficulty Metrics}—\textit{Fertility Ratio (F)}, \textit{Retrieval Proxy} ($R$), \textit{Pre-training Exposure} ($E$), and \textit{Corpus Diversity} ($D$)—to contextualize MT benchmarks and identify when high scores correlate with low diversity or contamination.

Our contributions are as follows:

\begin{itemize}[nosep,leftmargin=*]
    \item We provide a systematic analysis of data quality factors driving result variability in \textbf{XLR MT}, advancing understanding of when and why transfer learning from high-resource languages fails.
    \item We establish \textbf{high-resource reference baselines} by constraining data to XLR sizes ($10^3$-$10^6$ sentences), providing controlled performance references.
    \item We introduce the \textbf{FRED Difficulty Metrics} that quantify task complexity independent of model performance, improving transparency of reported gains.
    \item We identify that underperforming outliers suffer from poor lexicon overlap with pre-trained models, highlighting structural limits of transfer learning for non-digitized languages.
\end{itemize}

\section{Methods}
\label{ch:data_status}

XLR languages exhibit extreme performance variability (BLEU 5--40+) without interpretable context, making it impossible to distinguish genuine translation capability from benchmark artifacts. We address this through two complementary approaches: (1) establishing controlled reference baselines, and (2) introducing difficulty metrics that quantify dataset characteristics.

\subsection{Establishing High-Resource Reference Baselines}
\label{ch:high_res}

A key challenge in evaluating XLR performance is the lack of controlled baselines. To address this, we establish reference baselines by artificially restricting high-resource MT systems to extremely low-resource sizes (training data ranging from $10^3$ to $10^6$ sentences), providing a ``gold standard'' for expected behavior under data-constrained conditions.

We select five typologically diverse languages—Finnish (fi), Chinese (zh), Arabic (ar), Japanese (ja), and Hindi (hi)—with details shown in Appendix Table \ref{tb:high_res} and fine-tune mBART on them. These languages represent distinct language families and writing systems (Alphabetic, Logographic, Abjad, Moraic, and Abugida orthographies), ensuring the baselines account for different morphological and orthographic challenges independent of data availability. 

\subsection{FRED Difficulty Metric Definition}

The performance of MT systems is heavily influenced by the underlying data distribution and benchmark characteristics. To move beyond raw performance scores, we propose four \textbf{Difficulty Metrics} that quantify dataset complexity (Table \ref{tab:metric_summary}).

\begin{table}[t]
\centering
\resizebox{\linewidth}{!}{
\begin{tabular}{@{}lcp{7.5cm}@{}}
\toprule
\textbf{Metric} & \textbf{Notation} & \textbf{Interpretation} \\
\midrule
Fertility Ratio & $F = N_{\text{token}} / N_{\text{char}}$ & Tokenization efficiency; \textbf{higher = harder} \\
Retrieval Proxy & $R$ & Upper bound via memorization; \textbf{lower = harder} \\
Pre-training Exposure & $E$ & Overlap with pre-training corpus; \textbf{lower = harder} \\
Corpus Diversity & $D$ & Train-test lexical similarity; \textbf{lower = harder} \\
\bottomrule
\end{tabular}
}
\caption{Summary of the four core metrics ($F$, $R$, $E$, $D$) quantify intrinsic task difficulty.}
\label{tab:metric_summary}
\end{table}

We describe all metrics below, and refer to appendix \ref{ch:metric_imp_detail} for detailed implementation. 
By design, these metrics are computationally efficient and non-parametric, with the computational overhead reported in Appendix \ref{ch:computecost}.

\vspace{2pt}\noindent\textbf{Token Fertility ($F$)}, defined as the ratio of tokens to characters\footnote{For non-latin languages, the number of characters is counted by  \texttt{len(str.split())} in Python.} ($N_{\text{token}} / N_{\text{char}}$), which captures the efficiency of tokenization. We calculate the $F$-score on both the source and target sides and report the larger one. Note that this is not a commonly seen fertility metric in the literature, but we find it is a good indicator of tokenization quality. For example, for extinct languages, there is no unicode code point overlapped with modern pretraining data.

\vspace{2pt}\noindent\textbf{Retrieval Proxy ($R$).} 
The $R$-score simulates the performance of a perfect retrieval-based system, establishing a ceiling on what can be achieved through memorization alone. For each test sentence, we identify its nearest neighbor in the training set and measure target similarity:
\vspace{-0.3em}
\begin{equation}
    R(f) = \frac{1}{M} \sum_{i \in D_{\text{te}}} f(y_i, y_{j^*}), 
\end{equation}
where $j^* = \operatorname*{arg\,max}_{j: (x_j, y_j) \in D_{\text{tr}}} f(x_i, x_j)$, and $f$ denotes a base metric such as BLEU.
Higher $R$-scores indicate that simple nearest-neighbor retrieval without cross-lingual understanding can achieve competitive results, signaling low inherent task difficulty. We adopt $R$-score rather than training a full phrase-based SMT (PBSMT) pipeline because it is substantially more compute-efficient and requires far fewer hyperparameters, while remaining by design similar to SMT performance, as we show in Appendix~\ref{ch:smt_correlation}.

\vspace{2pt}\noindent\textbf{Pre-training Exposure ($E$)}, which quantifies overlap between evaluation data and the model's pre-training corpus. Let $G_{\text{te}}$ be the set of unique $n$-grams in the test set target sentences, and $\operatorname{count}(g, D_{\text{pt}})$ be the frequency of $4$-gram\footnote{Any $n$-gram size can be used, 4 is out of rule-of-thumb.} $g$ within the pre-training corpus $D_{\text{pt}}$, calculated using \textit{infini-gram} \cite{liu2025infinigramscalingunboundedngram}:
\vspace{-0.5em}
\begin{equation}
    E = \frac{1}{|G_{\text{te}}|} \sum_{g \in G_{\text{te}}} \operatorname{count}(g, D_{\text{pt}}).
\end{equation}
 A high $E$-score indicates that test data contains phrases frequently seen during pre-training, suggesting the model may rely on memorization rather than cross-lingual transfer.

\vspace{2pt}\noindent\textbf{Corpus Diversity ($D$).} 
This metric evaluates lexical diversity by measuring similarity between training and test instances, similar to self-BLEU \citep{Zhu2018TexygenAB}. We compute the average pairwise similarity between all training and test instances:
\vspace{-0.3em}
\begin{equation}
    D(f) = \frac{1}{NM} \sum_{i \in D_{\text{te}}} \sum_{j \in D_{\text{tr}}} f(y_i, y_j)
\end{equation}
A high $D$-score indicates similar vocabulary and phrasal patterns between training and test sets (low lexical diversity), which is a common phenomenon in domain-restricted corpora.

\begin{table}[h]
\centering
\resizebox{\linewidth}{!}{
\begin{tabular}{@{}cccccccc@{}}
\toprule
\multirow{2}{*}{\textbf{Lang}} & \multirow{2}{*}{\shortstack{\textbf{$N_{\text{train}}$}\\(\# sent)}} & \multirow{2}{*}{\textbf{$N_{\text{token}}$}} & \multirow{2}{*}{\textbf{$F$-score}} & \textbf{$E$-score} & \textbf{$D$-score} & \textbf{$R$-score} & \textbf{Reported scores} \\
\cmidrule{5-5}\cmidrule(lr){6-6} \cmidrule(lr){7-7} \cmidrule{8-8}
 & & & & 4-gram & BLEU & BLEU & BLEU \\
\midrule
\multicolumn{8}{l}{\textit{High-resource languages} (Table \ref{tab:high-resource-appendix})}\\
avg. & 10k & 30.8 & 0.41 & 96.9 & 1.37 & 3.24 & 16.27 \\
ja$\rightarrow$en  & 10k & 31.4 & 0.56 & 114 & 1.65  & 2.86 & 9.03 \\
hi$\rightarrow$en  & 10k & 33.1  & 0.28 & 86 & 0.95 & 2.78 & 14.02 \\
fi$\rightarrow$en  & 10k & 27.3 &0.25 & 85 & 2.03 & 3.21 & 13.46 \\
zh$\rightarrow$en  & 10k & 38.0 & 0.66 & 90 & 0.78 & 2.38 & 13.32 \\
ar$\rightarrow$en  & 10k & 24.0 & 0.32 & 109 & 1.45 & 4.98 & 31.54 \\
\midrule
\multicolumn{8}{l}{\textit{Ancient (extinct) languages}  
 (\citet{chen2024logogramNLP,cao-etal-2024-deep}, Table \ref{tab:ancient-appendix})} \\
akk$\rightarrow$en  & 50k & 24.6 & 1.00 & 82.2 & 1.59  &  32.10 & 44.41 \\
egy$\rightarrow$en/de  & 10k & 24.0 & 1.00 & 0.08 & 3.47 & 23.43 & 34.45 \\
\midrule
\multicolumn{8}{l}{\textit{Formosan languages (indigenous languages in Taiwan)} (\citet{zheng-etal-2024-improving-low}, Table \ref{tab:formosan-mandarin})} \\
avg. & - & 15.2 & 0.92 & 0.006 & 5.27 & 13.81 & 8.14 \\
tao$\rightarrow$zh & 5k & 11.0 & 0.91 & 0.08 & 5.34 & 17.80 & 14.74 \\
\midrule
\multicolumn{8}{l}{\textit{Americas Indigenous Languages} (\citet{de-gibert-etal-2025-findings}, Table \ref{tb:americas})} \\
avg.& - & 38.8 & 0.40 & 1.17 & 1.83 & 4.72 & 5.98 \\
& & & & &  (11.19) & (15.06) & (26.41) \\
shp$\rightarrow$es  & 14k & 20.9 & 0.33 & 0.65 & 3.83 & 4.86 & 7.22 \\
& & & &  & (8.69) & (14.43) & (27.33) \\
hch$\rightarrow$es & 9k & 26.9 & 0.40 & 0.65 & 3.05 & 3.41 & 3.69 \\
& & & & & (11.54) & (13.65) & (23.26) \\
quy$\rightarrow$es & 125k & 22.3 & 0.34 & 0.65 & 1.27 & 3.85 & 8.76 \\
& & & & & (13.77) & (12.79) & (33.83) \\
guc$\rightarrow$es & 59k & 35.9 & 0.40 & 0.24 & 0.93 & 8.15 & 2.22 \\
& & & & & (13.35) & (23.89) & (12.58) \\
\midrule
\multicolumn{8}{l}{\textit{African indigenous languages} (\citet{adelani2022few}, Table \ref{tab:african-nlp})} \\
avg. & - & 51.7 & 0.39 & 1.43 & 5.20 & 13.2 & 15.1 \\
hau$\rightarrow$en & 3k & 46.1 & 0.29 & 146 & 1.41 & 6.28 & 12.9 \\
zul$\rightarrow$en & 3k & 49.9 & 0.33 & 99.6 & 1.41 & 32.85 & 31.1 \\
bam$\rightarrow$fr & 3k & 56.2 & 0.45 & 2.65 & 1.79 & 7.28 & 10.0 \\
\midrule
\multicolumn{8}{l}{\textit{Indic indigenous Languages} (\citet{pal-etal-2023-findings}, Table \ref{tb:indic})} \\
mni$\rightarrow$en & 50k & 48.1 & 0.54 & 330 & 1.24 & 19.91 & 69.75 \\
kha$\rightarrow$en & 24k & 60.5 & 0.39 & 727 & 2.00 & 4.43 & 20.72 \\
\bottomrule
\end{tabular}
} 
\caption{Overview of FRED difficulty metrics to measure the data quality of different languages (translating into high-resource direction). The column of \textbf{Reported scores} of high-resource language group are shown here for reference and trained by us and the low-resource pairs from different languages groups are excerpted from corresponding papers (citations are listed in the table). $E$-score is calculated on the target side.
 \textbf{$^{*}$} For Americas Group, we also reported chrF++ score in parenthesis below BLEU score. For a complete data, refer to Appendix (Table \ref{tb:quality_check_into_lowres}).}
\label{tb:quality_check}
\end{table}

\section{Experiments and Analysis\label{ch:analysis}}

\subsection{Dataset Collection and Analysis}

We surveyed low-resource workshops and papers at *CL conferences over the past three years. As shown in Appendix Table \ref{tb:xlr_survey}, XLR languages fall into three groups: (1) under-represented languages with substantial speakers but limited digital presence (e.g., African and Indic); (2) endangered languages with small speaker communities (e.g., Formosan and Americas indigenous); and (3) ancient (extinct) languages with fixed corpora (e.g., Ancient Egyptian and Akkadian).

For fair comparison, we report numbers without extra training data, primarily using pre-trained models. Table \ref{tb:quality_check} shows metrics for translation into high-resource direction; the reverse direction is in Appendix Table \ref{tb:quality_check_into_lowres}.

\subsection{Interpreting Metrics Against Baselines}

The high-resource group is a synthetic setting where we cap training at 10k parallel sentences to approximate low data volume. This baseline cannot reproduce the full diversity of authentic low-resource conditions---e.g., orthographic and transcription practices, domain shift, and small-community data ecosystems---but it isolates how otherwise high-resource language pairs behave when the \emph{only} bottleneck is limited supervised training data.

\paragraph{High-resource baselines establish expected ranges.} 
It exhibits mean D-BLEU of 1.37 and R-BLEU of 3.24, indicating relatively high sentence diversity and low train-test overlap. The average $E$-score of 96.9 shows moderate pre-training exposure. \textbf{These values provide reference points}: XLR languages significantly deviating from these ranges likely have data quality issues rather than inherent linguistic difficulty.

\paragraph{Not all XLR languages can match baseline diversity.} 
For ancient languages like Akkadian (akk) and Ancient Egyptian (egy), $R$-scores of 32.10 and 23.43 far exceed the baseline range (3.24). \textbf{This is not an error but a real constraint}: with fixed, limited corpora, achieving low train-test similarity is infeasible. The low-resource setting is not realistically fixable for extinct languages, since no new native text can be produced and corpora cannot grow beyond what has been attested. However, these high $R$-scores explain their unexpectedly high BLEU scores (44.41 and 34.45)--the task is genuinely easier due to memorization opportunities.

\subsection{Metric Correlation Analysis}

\begin{table}[h]
    \centering
    \resizebox{\linewidth}{!}{
    \begin{tabular}{lcl}

    \hline
    \textbf{Feature} & \textbf{$R^2$ Value} & \textbf{Pred Strength} \\ \hline
    $R$-score & 0.5821 & Strongest \\
    $N_{token}$ & 0.3415 & Moderate \\
    $F$-score ($N_{token}/N_{char}$) & 0.2248 & Low-Moderate \\
    $D$-score & 0.1204 & Low \\
    $E$-score & 0.0142 & Negligible \\
    $N_{\text{train}}$ & 0.0011 & None \\ \hline
    \end{tabular}
    }
\caption{Individual $R^2$ for Hybrid Regression Model (BLEU/ChrF/chrF++).\label{tab:r2_scores}}
\end{table}

We conducted a regression correlation analysis between the proposed metrics, training size, and model performance (details can be found in Appendix \ref{ch:regression_corr}). $R$-score emerges as the dominant predictor ($R^2$ = 0.582), explaining over 58\% of performance variance. This confirms our hypothesis that \textbf{train-test relationships matter more than training size} in XLR MT. Additionally, we found $E$-score has very little correlation compared to other factors, which suggests in XLR MT, pre-training data contamination is not the most significant issue.

\subsection{Specific Findings for Language Groups}

\noindent\textbf{High performance explained by high $R$-scores.} 
Ancient languages such as Akkadian and Ancient Egyptian achieve BLEU scores exceeding 40—dramatically above the high-resource baseline (Figure \ref{fig:overview}). \textbf{Our metrics reveal the reason}: their $R$-scores (32.10 and 23.43) are 6-10× higher than the baseline (3.24), indicating that even perfect retrieval would achieve strong performance. These languages represent the easiest XLR translation tasks due to limited corpus diversity.

\vspace{2pt}\noindent\textbf{Token fertility reveals structural limits.}
Formosan and ancient languages show fertility approaching 1.0, meaning nearly every character becomes a separate token, \textbf{indicating a tokenization failure}: the pre-trained model lacks subword units for efficient representation. Compared to Americas languages with better tokenization (fertility ~0.40), these languages show marginal improvement over $R$-score baselines. Formosan neural models average BLEU 8.14, far below the $R$-score of 13.81, suggesting that these models cannot learn from poorly represented data.

\vspace{2pt}\noindent\textbf{Neural models should outperform retrieval baselines.} 
The $R$-score represents a trivial lower bound, i.e., what nearest-neighbor matching achieves without any learning. \textbf{Yet many XLR MT systems fail to surpass it}. For example, Zulu's $R$-score achieves BLEU 32.85, exceeding the best neural model (21.2). Similarly, some Americas languages, such as Wayuunaiki (guc), peak at ChrF++ 12.58, well below their $R$-score (23.89). This suggests systematic underfitting or poor hyperparameter tuning \cite{sennrich2019revisiting}, as effective and meaningful neural models should surpass simple retrieval.

\vspace{2pt}\noindent\textbf{Pre-training gaps explain Americas underperformance.} 
Despite comparable $D$-scores and $R$-scores to high-resource baselines, Americas languages achieve lower ChrF. Their $E$-scores (0.24-0.65) are 100$\times$ lower than high-resource languages (86-114). Without pre-training coverage, these languages rely on cross-lingual transfer from limited parallel data, validating prior observations about monolingual data importance \citep{mager-etal-2023-neural}.


\section{Conclusion}

We investigated variability in XLR machine translation performance, establishing that much stems from dataset characteristics rather than linguistic properties or model capabilities. Our analysis reveals clear patterns: overperforming outliers benefit from high train-test similarity or pre-training exposure, while underperforming outliers suffer from poor tokenization and minimal pre-training representation. These findings highlight fundamental limitations: pre-trained multilingual models cannot effectively transfer to languages outside their representation space.

We strongly encourage future XLR MT research to report the proposed \textbf{FRED Difficulty Metrics} alongside standard metrics (BLEU, ChrF, and etc.), enabling reliable cross-study comparisons and helping distinguish genuine methodological advances from benchmark artifacts.


\section*{Limitations}

While BLEU or ChrF (and their variants) are commonly used evaluation metrics, their comparability across different languages remains challenging. The automatic FRED difficulty metrics we propose are a step toward better evaluation for extremely low-resource languages, but there is still room for improvement, particularly in measuring lexicon overlap and token fertility.


The performance of mBART on the five high-resource languages, though informative, could be better characterized through more fine-grained analyses. Future work could also benefit from more detailed analyses to better identify outliers among XLR languages in machine translation.

We wish we had more time to investigate the correlation between monolingual data and model performance. We could also apply a similar $E$-score to monolingual data.

\section*{Ethics Statement}

This work highlights the challenges faced by extremely low-resource languages, which we define as languages with fewer than 1M training examples and no additional unlabeled resources. By emphasizing this definition, we aim to underscore the need for more effective cross-lingual transfer learning approaches that can operate in data-scarce scenarios.

We made efforts to ensure that the data examined in this paper represent languages from diverse regions around the world, thereby promoting inclusivity and comprehensiveness in our analysis.


\bibliography{anthology,custom}
\bibliographystyle{acl_natbib}

\clearpage

\appendix

\section*{Appendix}

\section{Regression Correlation Analysis\label{ch:regression_corr}}

We conducted a regression correlation analysis on the $X$ and $Y$ for language pairs listed in Table \ref{tb:quality_check}.

$X$: $N_{\text{train}}$, $N_{\text{token}}$, $F$-score, $R$-score, $E$-score, $D$-score

$Y$: neural models' performance in BLEU (except for native Americas languages, we use ChrF++ instead of BLEU, the corresponding RED scores are also using ChrF++)



\section{Tables}

\subsection{Low resource Machine Translation survey}

Table \ref{tb:xlr_survey} shows our survey of venues and publications on XLR languages from the past three years.

\begin{table*}[h]
\centering
\resizebox{0.7\linewidth}{!}{
\begin{tabular}{@{}cccc@{}}
\toprule
\textbf{Venue}    & \textbf{Language}  & \textbf{Region/Period}   & \textbf{Reference} \\ \midrule
WMT & Multiple & Global & \citet{kocmi-etal-2023-findings} \\
AfricaNLP & Multiple & African & \citet{georgetangalenlp} \\
AmericasNLP    &    Multiple      &    Latin America     &    \citet{ebrahimi-etal-2024-findings}       \\ 
WAT/WMT & Indic & South Asia & \citet{nakazawa-etal-2023-overview,pal-etal-2023-findings,zhou-etal-2025-transsionmts}\\
ML4AL & Ancient Egyptian & Ancient &  \citet{cao-etal-2024-deep} \\
*CL Conf & Akkadian & Ancient &  \citet{chen2024logogramNLP}\\
LoResMT & Cantonese & East Asia  & \citet{loresmt-2023-technologies}\\
*CL Conf &  Formosan & East Asia & \citet{zheng-etal-2024-improving-low}\\
*CL Conf & Northern Sámi & Europe & \citet{saleva-lignos-2024-language}\\
\bottomrule
\end{tabular}
} 
\caption{Survey of venues and publications on XLR languages from the past three years.}
\label{tb:xlr_survey}
\end{table*}

\subsection{High resource MT statistics}
Table \ref{tb:high_res} shows the High-resource reference baselines (xx→en). BLEU / ChrF scores with capped training data (number of
training sentences). Languages are selected from varied language families and writing systems to represent diverse
morphological and orthographic challenges.

\begin{table*}[h]
    \centering
    \resizebox{0.8\textwidth}{!}{
    \begin{tabular}{ccccccccc}
    \toprule
    \multirow{2}{*}{\textbf{Lang}} & \multirow{2}{*}{\textbf{Writing}} & \multirow{2}{*}{\textbf{Phonography}} & \multirow{2}{*}{\textbf{Dataset}} & \multicolumn{4}{c}{\textbf{BLEU / ChrF at Different Data Sizes}} \\
    \cmidrule(lr){5-8}
    & & & & \textbf{1k} & \textbf{10k} & \textbf{100k} & \textbf{1M} \\
    \midrule
    fi & Latin & Alphabetic & wmt18-fi-en & 6.32 / 28.87 & 11.97 / 38.82 & 18.31 / 45.39 & 18.47 / 46.98 \\
    zh & Han & Logographic & wmt18-zh-en & 6.51 / 30.05 & 10.77 / 38.10 & 15.42 / 44.67 & 17.59 / 47.34 \\
    ar & Arabic & Abjad & iwslt2017-ar-en & 7.20 / 24.49 & 18.94 / 38.90 & 29.61 / 50.10 & 36.31 / 56.03 \\
    ja & Han/Kana & Moraic & iwslt2017-ja-en & 3.08 / 18.97 & 7.17 / 31.37 & 11.24 / 37.49 & -- \\
    hi & Devanagari & Abugida & IITB-hi-en & 1.42 / 15.88 & 4.40 / 26.47 & 7.71 / 39.17 & 8.11 / 45.66 \\
    \midrule
    \multicolumn{4}{c}{\textbf{avg.}} & 4.91 / 23.65 & 10.65 / 34.73 & 16.46 / 43.36 & 20.12 / 49.01 \\
    \multicolumn{4}{c}{\textbf{std.}} & 2.25 / 5.50 & 4.93 / 5.00 & 7.50 / 4.54 & 10.19 / 4.11 \\
    \bottomrule
    \end{tabular}
    }
    \caption{High-resource reference baselines (xx$\rightarrow$en). BLEU / ChrF scores with capped training data (number of training sentences). Languages are selected from varied language families and writing systems to represent diverse morphological and orthographic challenges.}
    \label{tb:high_res}
\end{table*}

\section{Full Overview of Automatic Metrics}
Table \ref{tb:quality_check_into_lowres} shows the overview of automatic metrics to measure the data quality of different languages from both low-res to hi-res and hi-res to low-res directions.

\begin{table*}[h]
\centering
\resizebox{\linewidth}{!}{
\begin{tabular}{@{}ccccccccccccc@{}} 
\toprule
\multirow{2}{*}{\textbf{Lang}} & \multirow{2}{*}{\shortstack{\textbf{Parallel}\\(\# sent)}} &  \multirow{2}{*}{\textbf{Monolingual}} & \multirow{2}{*}{\textbf{$N_{\text{token}}$}} & \multirow{2}{*}{\textbf{$N_{\text{char}}$}} & \multirow{2}{*}{\textbf{$\frac{N_{\text{token}}}{N_{\text{char}}}$}} & \textbf{$D$-score} & \textbf{$R$-score} & \textbf{PBSMT} & \textbf{$E$-score} & \textbf{Model} \\
\cmidrule(lr){7-7} \cmidrule(lr){8-8} \cmidrule(lr){9-9} \cmidrule(lr){10-10} \cmidrule(lr){11-11}
 & & & & & & \textbf{BLEU / ChrF} & \textbf{BLEU / ChrF} & \textbf{BLEU / ChrF}  & \textbf{4-gram} & \textbf{BLEU / ChrF} \\
\midrule

\multicolumn{12}{l}{\textbf{High-resource languages} (Table \ref{tab:high-resource-appendix})} \\
ja$\rightarrow$en  & 10k & >10M & 31.4 & 56.3 & 0.56 & 1.65 / 5.01  & 2.86 / 18.38 & 4.49 / 21.52 & 114.47 & 9.03 / 31.37  \\
hi$\rightarrow$en  & 10k & >10M& 33.1 & 117.2 & 0.28 & 0.95 / 10.08 & 2.78 / 17.34 & 2.36 / 11.20 & 85.98 & 14.02 / 26.47 \\
fi$\rightarrow$en  & 10k & >10M& 27.3 & 109.8 & 0.25 & 2.03 / 16.44 & 3.21 / 17.16 & 4.63 / 24.01 & 85.46 & 13.46 / 38.82 \\
zh$\rightarrow$en  & 10k & >10M& 38.0 & 59.0 & 0.66 & 0.78 / 1.60 & 2.38 / 17.46 & 4.18 / 24.87 & 90.17 & 13.32 / 38.10  \\
ar$\rightarrow$en  & 10k & >10M & 24.0 & 75.7 & 0.32 & 1.45 / 11.36 & 4.98 / 18.56 & 8.34 / 22.71 & 108.57 & 31.54 / 38.90 \\
\\
en$\rightarrow$ja  & 10k & >10M & 31.0 & 129.9 & 0.24 & 1.75 / 16.84  & 2.88 / 6.46 & 5.62 / 9.46 & 0.08 & -  \\
en$\rightarrow$hi  & 10k & >10M& 27.6  & 117.6 & 0.24 & 1.39 / 12.60 & 2.28 / 14.92 & 2.11 / 6.95 & 20.7 & - \\
en$\rightarrow$fi  & 10k & >10M& 25.8 & 108.9 & 0.24 & 1.66 / 15.63 & 3.74 / 18.06 & 4.47 / 27.10 & 0.27 & - \\
en$\rightarrow$zh  & 10k & >10M& 37.2 & 156.5 & 0.24 & 1.30 / 15.25 & 1.95 / 3.09 & 8.89 / 9.82 & 0.56 & - \\
en$\rightarrow$ar  & 10k & >10M & 22.8 & 94.0 & 0.24 & 2.01 / 13.64 & 3.58 / 15.09 & 7.34 / 23.22 & 0.25 & - \\
\midrule

\multicolumn{12}{l}{\textbf{Ancient (extinct) languages}  
 (\citet{chen2024logogramNLP,cao-etal-2024-deep}, Table \ref{tab:ancient-appendix})} \\
akk$\rightarrow$en  & 50k & 0 & 24.6 & 24.6 & \underline{1.00} & 1.59 / 2.61 &  32.10 / 46.18 & 23.86 / 43.25 & 82.21
 & 44.41 \\
egy$\rightarrow$en/de  & 10k & 0 & 24.0 & 24.0 & \underline{1.00} & 3.47 / 9.22 & 23.43 / 41.95 & 3.39 / 13.86
 & 0.08 & 34.45 &\\
\\
en$\rightarrow$akk  & 50k & 0 & 26.2 & 86.9 & 0.30 & 1.60 / 8.50  & 29.79 / 32.37  & 31.81 / 33.08  & 0 & - \\
en/de$\rightarrow$egy  & 10k & 0 & 18.8 & 63.6 & 0.30 & 2.15 / 11.92 & 31.08 / 41.64 & 6.69 / 13.83
 & 1.12 & - \\
\midrule
\multicolumn{12}{l}{\textbf{Formosan languages (indigenous languages in Taiwan)} (\citet{zheng-etal-2024-improving-low}, Table \ref{tab:formosan-mandarin})} \\
tao$\rightarrow$zh & 5k & 0 & 11.0 & 30.8 & 0.36 & 5.34 / 31.06 & 17.80 / 18.47 & 11.22 / 9.19 & 0.08 & 4.72 \\
\\
zh$\rightarrow$tao & 5k & 0 & 10.6 & 11.7 & \underline{0.91} & 3.41 / 2.45 & 19.22 / 34.84 & 11.73 / 17.86 & 0.0004 & 20.32\\
\multicolumn{12}{l}{\textbf{Americas Indigenous Languages} (\citet{de-gibert-etal-2025-findings}, \citet{yahan-islam-2025-leveraging}, Table \ref{tb:americas})} \\
shp$\rightarrow$es  & 14k & 0 & 20.9 & 64.4 & 0.33 & 3.98 / 8.69 & 5.42 / 14.43 & 7.30 / 21.11 & 0.65 & 7.22 / 27.33 & \\
hch$\rightarrow$es & 9k & 0 & 26.9 & 66.9 & 0.40 & 3.18 / 11.54 & 3.91 / 13.65 & 5.65 / 18.20 & 0.65 & 3.69 / 23.26 \\ 
quy$\rightarrow$es & 125k & - & 22.3 & 65.7 & 0.34  & 2.60 / 13.77 & 4.66 / 12.79 & 7.51 / 20.60 & 0.65  & 8.76 / 33.83 \\
guc$\rightarrow$es & 59k & - & 35.9 & 35.9 & 0.40 & 0.93 / 13.35 &  8.15 / 23.89 &  9.42 / 23.89 & 0.24 & 2.22 / 12.58 \\
\\
es$\rightarrow$shp  & 14k & 0 & 15.1 & 64.5 & 0.23 & 2.77 / 9.19 & 5.71 / 17.98 & 5.95 / 22.49 & 0.14& 1.30 / 18.12 & \\
es$\rightarrow$hch & 9k & 0 & 15.2 & 64.7 & 0.23 & 1.87 / 11.08 & 4.72 / 19.90 & 6.90 / 23.66 & 0.37 & 8.66 / 28.17 \\ 
es$\rightarrow$quy & 125k & - & 15.1 & 64.5 & 0.23 & 1.94 / 13.68 & 3.37 / 19.05 & 5.40 / 29.08 & 1.05 & 2.43 / 40.01 \\
es$\rightarrow$guc & 59k & - & 18.0 & 76.2 & 0.24 & 1.22 / 13.62 & 8.36 / 31.93 & 5.24 / 28.19 & 0.15 & 1.11 / 17.56 \\
\midrule
\multicolumn{12}{l}{\textbf{African indigenous languages} (\citet{adelani2022few}, Table \ref{tab:african-nlp})} \\
hau$\rightarrow$en & 3k & 236k & 46.1 & 159.8 & 0.29 & 1.41 / 18.82 & 6.28 / 23.39 & 6.09 / 29.69 & 146 & 12.9 \\
zul$\rightarrow$en & 3k  & 667k & 50.0 & 153.3 & 0.33 & 1.41 / 16.67 & 32.85 / 47.29 & 2.47 / 15.97 & 99.6 & 31.1\\
bam$\rightarrow$fr & 3k & - & 56.2 & 123.7 & 0.45 & 1.79 / 14.76 & 7.28 / 22.46 & 6.62 / 27.78 & 2.65 & 10.0 \\
\\
en$\rightarrow$hau & 3k & 236k & 31.2 & 138.9 & 0.22 & 1.64 / 16.10 & 5.56 / 26.20 & 5.54 / 31.54 & 0.10 & 10.4 \\
en$\rightarrow$zul & 3k  & 667k & 33.2 & 136.8 & 0.24 & 1.34 / 14.72 & 11.55 / 34.94 & 2.39 / 15.26 & 2.81 & 21.2  \\
fr$\rightarrow$bam & 3k & - & 34.4 & 131.2 & 0.26 & 1.74 / 15.08 & 6.35 / 23.12 & 7.09 / 27.88 & 1.71 & 18.6 \\
\midrule
\multicolumn{12}{l}{\textbf{Indic indigenous Languages} (\citet{pal-etal-2023-findings}, Table \ref{tb:indic})} \\
mni$\rightarrow$en & 50k & 4M & 48.2 & 89.8 & 0.54 & 1.24 / 13.07 & 19.91 / 38.84 & 14.85 / 42.69 & 330 & 69.75 \\
kha$\rightarrow$en & 24k & 910k & 60.5 & 157.0 & 0.39 & 2.00 / 19.61 & 4.43 / 19.51 & 7.67 / 31.82 & 727 & 20.72  \\
\\
en$\rightarrow$mni & 50k & 4M & 19.4 & 89.0 & 0.22 & 1.57 / 14.87 & 17.80 / 37.66 & 4.86 / 43.61 & 5.79 & 29.50 \\
en$\rightarrow$kha & 24k & 910k & 31.3 & 113.8 & 0.28 & 1.93 / 17.71 & 5.39 / 25.41 & 11.17 / 36.32 & 2.94 & 21.63 \\
\bottomrule
\end{tabular}
}
\caption{Overview of automatic metrics to measure the data quality of different languages on both low-to-high and high-to-low directions. The high-resource language (ja, hi, fi, zh, ar) are shown here for reference. $N_{\text{token}}$, $N_{\text{char}}$ and $\frac{N_{\text{token}}}{N_{\text{char}}}$ are calculated on the test dataset of the source side, TTR is calculated on the train dataset of the target side, and $E$-score is calculated on the test dataset on the target side. For Americas languages, ChrF++ scores are reported instead of ChrF in Model column.}
\label{tb:quality_check_into_lowres}
\end{table*}

\section{Tables in details for different MT benchmark}
\label{sec:appendix}

\begin{table}[h]
\centering
\resizebox{\linewidth}{!}{
\begin{tabular}{@{}lcccc@{}}
\toprule
\textbf{Lang} & \textbf{N-train} & \textbf{D-chrF} & \textbf{R-chrF} & \textbf{PBSMT-chrF} \\ 
\midrule
Japanese (ja) & 10000 & 5.01 & 18.38 & 21.52  \\
Hindi (hi) & 10000 & 10.08 & 17.34 & 11.20 \\
Finnish (fi) & 10000 & 16.44 & 17.16 & 24.01 \\
Chinese (zh) & 10000 & 1.60 & 17.46 & 24.87 \\
Arabic (ar) & 10000 & 11.36 & 18.56 & 22.71 \\
\midrule
\textbf{Lang} &  & \textbf{D-chrF++} & \textbf{R-chrF++} & \textbf{PBSMT-chrF++}  \\ 
\midrule
Japanese (ja) &  & 3.76 & 16.35 & 19.71  \\
Hindi (hi) &  & 8.37 & 15.10 & 11.04  \\
Finnish (fi) &  & 13.66 & 15.26 & 21.38  \\
Chinese (zh) &  & 1.32 & 13.80 & 22.42  \\
Arabic (ar) &  & 9.30 & 17.11 & 22.48  \\
\bottomrule
\end{tabular}
}
\caption{Translation Performance Metrics for high-resource language pairs -- chrF as similarity function.}
\label{tab:high-resource-appendix}
\end{table}

\begin{table}[h]
\centering
\resizebox{\linewidth}{!}{
\begin{tabular}{@{}lcccc@{}}
\toprule
\textbf{Lang} & \textbf{N-train} & \textbf{D-chrF} & \textbf{R-chrF} & \textbf{PBSMT-chrF}  \\ 
\midrule
Akkadian (akk) & 50000 & 2.61 & 46.18 & 43.25  \\
Egyptian (egy) & 10000 & 9.22 & 41.95 & 13.86 \\
\midrule
\textbf{Lang} &  & \textbf{D-chrF++} & \textbf{R-chrF++} & \textbf{PBSMT-chrF++}  \\ 
\midrule
Akkadian (akk) &  & 2.21 & 45.10 & 41.48  \\
Egyptian (egy) &  & 9.02 & 41.34 & 12.12 \\
\bottomrule
\end{tabular}
}
\caption{Translation Performance Metrics -- chrF and chrF++ for Ancient language pairs.}
\label{tab:ancient-appendix}
\end{table}

\begin{table}[h]
\centering
\resizebox{\linewidth}{!}{
\begin{tabular}{@{}lcccccc@{}}
\toprule
\textbf{Lang} & \textbf{N-train} & \textbf{D-BLEU} & \textbf{R-BLEU} & \textbf{PBSMT-BLEU} & \textbf{E-4-gram} & \textbf{BLEU} \\ 
\midrule
Sakizaya (ais) & 4590 & 6.09 & 12.05 & 9.17 & 0 & 3.11 \\
Amis (ami) & 4600 & 5.51 & 12.59 & 8.97 & 0 & 3.56 \\
Bunun (bnn) & 7180 & 6.21 & 15.7 & 13.97 & 0.002 & 5.44 \\
Kavalan (ckv) & 6573 & 5.04 & 17.61 & 19.81 & 0 & 7.18 \\
Rukai (dru) & 8319 & 5.82 & 10.48 & 11.39 & 0 & 8.44 \\
Paiwan (pwn) & 4126 & 5.24 & 9.93 & 9.77 & 0.001 & 3.80 \\
Puyuma (pyu) & 5515 & 4.12 & 10.23 & 11.22 & 0.001 & 7.86 \\
Seediq (sdq) & 4367 & 3.20 & 9.53 & 8.03 & 0 & 1.52 \\
Thao (ssf) & 5952 & 4.53 & 23.77 & 15.70 & 0 & 10.50 \\
Saaroa (sxr) & 3839 & 7.87 & 13.88 & 8.41 & 0 & 6.03 \\
Yami (tao) & 5186 & 5.34 & 17.8 & 11.22 & 0.082 & 4.72 \\
Atayal (tay) & 4600 & 5.51 & 12.59 & 8.96 & 0 & 4.86 \\
Truku (trv) & 3678 & 6.27 & 7.99 & 9.74 & 0.003 & 1.26 \\
Tsou (tsu) & 3550 & 5.62 & 17.61 & 9.57 & 0 & 2.07 \\
Kanakanavu (xnb) & 5294 & 3.68 & 16.56 & 16.57 & 0 & 9.54 \\
Saisiyat (xsy) & 4839 & 4.25 & 12.68 & 11.72 & 0 & 3.99 \\
\midrule
\textbf{Lang} &  & \textbf{D-chrF} & \textbf{R-chrF} & \textbf{PBSMT-chrF} &  & \textbf{chrF} \\ 
\midrule
Sakizaya (ais) & & 14.52 & 14.27 & 8.17 & & 14.38 \\
Amis (ami) & & 14.79 & 12.63 & 6.38 & & 12.08 \\
Bunun (bnn) &  & 16.18 & 16.74 & 10.60 & & 17.91 \\
Kavalan (ckv) &  & 15.09 & 18.88 & 16.03 &  & 24.03 \\
Rukai (dru) &  & 16.01 & 11.74 & 7.79 & & 36.66 \\
Paiwan (pwn) &  & 15.89 & 10.79 & 9.82 &  & 4.86 \\
Puyuma (pyu) & & 16.34 & 10.52 & 10.15 & & 15.69 \\
Seediq (sdq) &  & 13.11 & 12.24 & 8.61 &  & 13.24 \\
Thao (ssf) &  & 15.27 & 26.81 & 12.87 &  & 26.66 \\
Saaroa (sxr) &  & 17.55 & 14.60 & 6.97 &  & 14.06 \\
Yami (tao) &  & 13.06 & 18.47 & 9.19 &  & 18.27 \\
Atayal (tay) & & 14.79 & 12.63 & 6.39 &  & 12.26 \\
Truku (trv) &  & 11.82 & 7.92 & 7.95 & & 6.87 \\
Tsou (tsu) &  & 13.23 & 22.73 & 9.27 & & 19.50 \\
Kanakanavu (xnb) &  & 15.58 & 18.17 & 13.92 & & 20.93 \\
Saisiyat (xsy) &  & 15.15 & 13.02 & 10.40 & & 16.07 \\
\midrule
\textbf{Lang} &  & \textbf{D-chrF++} & \textbf{R-chrF++} & \textbf{PBSMT-chrF++} &  &  \\ 
\midrule
Sakizaya (ais) & & 13.58 & 12.23 & 10.57 & & \\
Amis (ami) & & 13.92 & 11.18 & 9.15 & & \\
Bunun (bnn) &  & 15.16 & 15.24 & 14.42 & & \\
Kavalan (ckv) &  & 13.58 & 17.00 & 20.82 &  & \\
Rukai (dru) &  & 14.32 & 10.26 & 11.36 & & \\
Paiwan (pwn) &  & 14.53 & 9.95 & 12.82 &  & \\
Puyuma (pyu) & & 14.39 & 9.59 & 13.81 & &\\
Seediq (sdq) &  & 11.58 & 11.10 & 10.86 &  & \\
Thao (ssf) &  & 13.66 & 24.54 & 17.10 &  & \\
Saaroa (sxr) &  & 15.44 & 13.89 & 9.74 &  & \\
Yami (tao) &  & 12.23 & 17.10 & 12.55 &  & \\
Atayal (tay) & & 13.92 & 11.18 & 9.15 &  &  \\
Truku (trv) &  & 11.69 & 7.10 & 10.72 & & \\
Tsou (tsu) &  & 13.04 & 20.48 & 11.48 & & \\
Kanakanavu (xnb) &  & 13.55 & 17.12 & 18.11 & & \\
Saisiyat (xsy) &  & 14.32 & 11.53 & 13.90 & & \\
\bottomrule
\end{tabular}
}
\caption{Translation Performance Metrics for Formosan-Chinese (Mandarin) language pairs. Exposure scores are calculated by counting on the high-resource side of the language pairs. The BLEU and chrF scores are taken from \citet{zheng-etal-2024-improving-low}.}
\label{tab:formosan-mandarin}
\end{table}

\begin{table}[h]
\centering
\resizebox{\linewidth}{!}{
\begin{tabular}{@{}lcccccc@{}}
\toprule
\textbf{lang} & \textbf{N-train} & \textbf{D-BLEU} & \textbf{R-BLEU} & \textbf{PBSMT-BLEU} & \textbf{E-4-gram} & \textbf{BLEU} \\ \midrule
ashaninka (cni) & 3883 & 3.98 & 4.94 & 6.36 & 0.62 & 2.35 \\ 
awajun (agr) & 21964 & 1.23 & 5.50 & 5.75 & 0.87 & 11.12 \\
aymara (aym) & 6531 & 1.74 & 3.51 & 5.17 & 0.65 & 8.82 \\ 
bribri (bzd) & 7508 & 2.01 & 4.33 & 6.01 & 0.65 & 4.31 \\ 
chatino (ctp) & 357 & 1.50 & 9.63 & 4.69 & 7.62 & - \\
guarani (gn) & 26032 & 1.89 & 4.51 & 6.29 & 0.65 & 8.62 \\ 
nahuatl (nah) & 16145 & 1.21 & 4.18 & 5.72 & 0.62 & 7.22 \\ 
otomi (oto) & 4889 & 0.63 & 2.47 & 3.78 & 0.67 & 1.50 \\ 
quechua (quy) & 125008 & 1.27 & 3.85 & 5.83 & 0.65 & 8.76 \\ 
raramuri (tar) & 14720 & 0.57 & 2.02 & 4.06 & 0.65 & - \\ 
shipibo (shp) & 14592 & 3.83 & 4.86 & 6.18 & 0.65 & 7.22 \\ 
wayuu (guc) & 59715 & 0.93 & 8.15 & 9.42 &  0.24 & 2.22 \\
wixarika (hch) & 8966 & 3.05 & 3.41 & 5.34 & 0.65 & 3.69 \\
\midrule
\textbf{lang} & & \textbf{D-chrF} & \textbf{R-chrF} & \textbf{PBSMT-chrF} &  &  \\ \midrule
ashaninka (cni) &  & 16.21 & 12.56 & 16.46 &  &  \\ 
awajun (agr) &   & 16.30  & 21.34 & 20.78 &  &  \\
aymara (aym) &  & 15.32 & 13.28 & 15.51 &  &  \\ 
bribri (bzd) &   & 6.37 & 16.03 & 20.40 &  & \\ 
chatino (ctp) &  & 19.90 & 29.41 & 25.63 & &  \\
guarani (gn) &  & 12.61 & 14.43 & 21.00 &  &  \\ 
nahuatl (nah) &  & 13.50 & 14.02 & 16.96 &  &  \\ 
otomi (oto) &  & 8.75 & 12.10 & 14.02 &  &  \\ 
quechua (quy) &  & 17.55 & 14.55 & 23.32 &  &  \\ 
raramuri (tar) &  & 7.82 & 13.92 & 16.42 &  & \\ 
shipibo (shp) & & 9.83 & 15.64 & 23.58 &  &  \\ 
wayuu (guc) &  & 17.04 & 26.33 & 25.66 &  &  \\
wixarika (hch) &  & 13.91 & 15.95 & 20.19 & & \\ 
\midrule
\textbf{lang} & & \textbf{D-chrF++} & \textbf{R-chrF++} & \textbf{PBSMT-chrF++} &  & \textbf{chrF++} \\ \midrule
ashaninka (cni) &  & 13.52 & 11.39 & 15.21 &  & 24.24 \\ 
awajun (agr) &   & 13.26  & 19.60 & 19.15 &  & 32.80 \\
aymara (aym) &  & 12.49 & 11.90 & 14.29 &  & 31.72 \\ 
bribri (bzd) &   & 6.54 & 13.72 & 18.32 &  & 26.74 \\ 
chatino (ctp) &  & 17.68 & 27.18 & 23.81 & & - \\
guarani (gn) &  & 10.70 & 12.68 & 19.32 &  & 32.07 \\ 
nahuatl (nah) &  & 10.77 & 12.51 & 15.28 &  & 26.89 \\ 
otomi (oto) &  & 6.93 & 10.42 & 12.12 &  & 19.01 \\ 
quechua (quy) &  & 13.77 & 12.79 & 20.60 &  & 33.83 \\ 
raramuri (tar) &  & 6.19 & 11.56 & 14.65 &  & - \\ 
shipibo (shp) & & 8.69 & 14.43 & 21.11 &  & 27.33 \\ 
wayuu (guc) &  & 13.35 & 23.89 & 23.89 &  & 12.58 \\
wixarika (hch) &  & 11.54 & 13.65 & 18.20 & & 23.26 \\ 
\bottomrule
\end{tabular}
}
\caption{Translation Performance Metrics for Americas Indigenous Languages - Spanish Language Pairs from 2025's shared tasks. We use dev set as the test set for the evaluation of this corpus since the test set of the dataset this year is not publicly released. Exposure scores are calculated by counting on the high-resource side of the language pairs. The BLEU and chrF++ scores are taken from the dev set performance of the team Syntax Squad in 2025's competition \cite{yahan-islam-2025-leveraging}. \label{tb:americas} }
\label{tab:spanish_language_pairs}
\end{table}

\begin{table}
\centering
\resizebox{\linewidth}{!}{
\begin{tabular}{@{}lcccccc@{}}
\toprule
\textbf{Lang} & \textbf{N-train} & \textbf{D-BLEU} & \textbf{R-BLEU} & \textbf{PBSMT-BLEU} & \textbf{E-4-gram} & \textbf{BLEU} \\ 
\midrule
\multicolumn{3}{l}{~~\textit{Translate into English}} & \\
Amharic (amh) & 899 & 0.29 & 2.80 & 4.42 & 109.65 & - \\
Hausa (hau) & 3098 & 1.41 & 6.28 & 6.09 & 146.42 & 12.9 \\
Igbo (ibo) & 6998 & 1.14 & 5.47 & 7.76 & 86.69 & 21.0 \\
Kinyarwanda (kin) & 460 & 1.24 & 2.95 & 4.24 & 94.52 & -  \\
Luganda (lug) & 4075 & 1.96 & 4.96 & 5.36 & 84.58 & 19.8 \\
Luo (luo) & 4262 & 1.50 & 2.99 & 2.54 & 79.44 & 12.1 \\
Chichewa (nya) & 483 & 1.15 & 2.40 & 2.83 & 91.70 & - \\
Nigerian-Pidgin (pcm) & 4790 & 1.47 & 4.33 & 18.83 & 96.85 & 44.2 \\
Shona (sna) & 556 & 1.27 & 2.87 & 3.30 & 101.49 & - \\
Swahili (swa) & 30782 & 1.38 & 5.72 & 11.98 & 91.70 & 29.5 \\
Setswana (tsn) & 2100 & 1.75 & 3.63 & 3.69 & 107.03 & 18.6 \\
Twi (twi) & 3337 & 1.49 & 2.37 & 3.31 & 98.85 & 9.8 \\
Xhosa (xho) & 486 & 1.86 & 3.52 & 4.07 & 107.90 & - \\
Yoruba (yor) & 6644 & 1.26 & 3.50 & 4.48 & 88.26 & 12.3 \\
Zulu (zul) & 3500 & 1.41 & 34.85 & 2.47 & 99.61 & 31.1 \\
\multicolumn{3}{l}{~~\textit{Translate into French}} & \\
Bambara (bam) & 3013 & 1.79 & 7.28 & 6.62 & 2.65 & 10.0 \\
Ghomala (bbj) & 2232 & 1.05 & 2.93 & 1.42 & 2.51 & 2.7 \\
Ewe (ewe) & 2026 & 2.12 & 1.92 & 2.87 & 4.55 & 4.1 \\
Fon (fon) & 2637 & 1.49 & 2.42 & 2.64 & 3.75 & 4.9 \\
Mossi (mos) & 2493 & 1.56 & 2.81 & 2.88 & 3.06 & 1.5 \\
Wolof (wol) & 3360 & 1.53 & 3.37 & 4.04 & 3.05 & 7.2 \\
\midrule
\textbf{Lang} & & \textbf{D-chrF} & \textbf{R-chrF} & \textbf{PBSMT-chrF} &  & chrF \\ 
\midrule
\multicolumn{3}{l}{~~\textit{Translate into English}} & \\
Amharic (amh) &  & 6.53 & 14.95 & 10.34 &  & - \\
Hausa (hau) &  & 18.82 & 23.39 & 29.69 &  & 33.2 \\
Igbo (ibo) &  & 12.86 & 22.59 & 32.41 &  & 46.4 \\
Kinyarwanda (kin) &  & 19.37 & 19.19 & 25.83 &  & -  \\
Luganda (lug) &  & 18.15 & 22.93 & 30.47 & & 45.4 \\
Luo (luo) &  & 18.83 & 19.46 & 22.66 &  & 34.1 \\
Chichewa (nya) &  & 20.63 & 18.57 & 23.93 &  & - \\
Nigerian-Pidgin (pcm) &  & 14.89 & 22.25 & 55.03 &  & 66.9 \\
Shona (sna) &  & 19.87 & 17.96 & 23.10 &  & - \\
Swahili (swa) &  & 15.44 & 22.11 & 40.83 &  & 53.7 \\
Setswana (tsn) &  & 18.84 & 20.00 & 25.00 & & 42.4 \\
Twi (twi) &  & 13.98 & 19.69 & 25.89 &  & 32.9 \\
Xhosa (xho) &  & 18.06 & 19.03 & 22.50 &  & - \\
Yoruba (yor) &  & 11.85 & 19.33 & 25.70 &  & 31.4 \\
Zulu (zul) &  & 16.67 & 47.29 & 15.97 &  & 43.9 \\
\multicolumn{3}{l}{~~\textit{Translate into French}} & \\
Bambara (bam) &  & 14.76 & 22.46 & 27.78 &  & 31.2 \\
Ghomala (bbj) &  & 11.01 & 19.12 & 16.98 &  & 21.8 \\
Ewe (ewe) &  & 12.34 & 17.28 & 21.16 &  & 24.8 \\
Fon (fon) &  & 11.84 & 19.38 & 20.82 &  & 20.5 \\
Mossi (mos) &  & 12.29 & 18.18 & 15.21 &  & 15.4 \\
Wolof (wol) &  & 13.88 & 19.61 & 24.77 &  & 26.2 \\
\midrule
\textbf{Lang} & & \textbf{D-chrF++} & \textbf{R-chrF++} & \textbf{PBSMT-chrF++} &  &  \\ 
\midrule
\multicolumn{3}{l}{~~\textit{Translate into English}} & \\
Amharic (amh) &  & 5.04 & 13.13 & 9.92 &  &  \\
Hausa (hau) &  & 16.03 & 21.30 & 27.57 &  &  \\
Igbo (ibo) &  & 10.66 & 20.25 & 28.94 &  &  \\
Kinyarwanda (kin) &  & 16.67 & 19.19 & 23.06 &  &   \\
Luganda (lug) &  & 14.82 & 20.70 & 26.96 & &  \\
Luo (luo) &  & 15.91 & 17.07 & 19.54 &  &  \\
Chichewa (nya) &  & 16.84 & 16.22 & 20.72 &  &  \\
Nigerian-Pidgin (pcm) &  & 12.86 & 19.35 & 50.67 &  & \\
Shona (sna) &  & 15.81 & 15.65 & 19.82 &  &  \\
Swahili (swa) &  & 12.90 & 19.72 & 37.62 &  &  \\
Setswana (tsn) &  & 16.47 & 17.69 & 22.30 & &  \\
Twi (twi) &  & 12.30 & 17.20 & 22.85 &  &  \\
Xhosa (xho) &  & 14.67 & 16.61 & 19.83 &  &  \\
Yoruba (yor) &  & 10.27 & 17.03 & 23.40 &  & \\
Zulu (zul) &  & 13.39 & 46.32 & 13.15 &  &  \\
\multicolumn{3}{l}{~~\textit{Translate into French}} & \\
Bambara (bam) &  & 13.20 & 20.35 & 25.62 &  &  \\
Ghomala (bbj) &  & 8.95 & 16.08 & 14.06 &  &  \\
Ewe (ewe) &  & 10.82 & 14.67 & 18.84 &  & \\
Fon (fon) &  & 10.84 & 16.83 & 18.96 &  & \\
Mossi (mos) &  & 10.67 & 15.93 & 13.64 &  & \\
Wolof (wol) &  & 12.08 & 16.90 & 22.40 &  & \\
\bottomrule
\end{tabular}
}
\caption{Translation Performance Metrics for African language pairs. Exposure scores are calculated by counting on the high-resource side of the language pairs. BLEU and chrF scores are taken from the best possible BLEU scores from mBART/mT5/T5 of table 4 of the paper  \cite{adelani2022thousandtranslationslongway}.}
\label{tab:african-nlp}
\end{table}

\subsection{Metrics Details}

For results of Table \ref{tab:high-resource-appendix}, \ref{tab:ancient-appendix}, \ref{tab:formosan-mandarin}, \ref{tab:spanish_language_pairs}, \ref{tab:african-nlp} and \ref{tab:indic-languages}, we report $N_{train}$ by directly counting number of training line pairs in the datasets. $D$-scores and $R$-scores are calculated on translating into high-resource direction. $E$-scores are calculated by counting on the high-resource side of the language pairs.

\begin{table}
\centering
\resizebox{\linewidth}{!}{
\begin{tabular}{@{}lcccccc@{}}
\toprule
\textbf{Lang} & \textbf{N-train} & \textbf{D-BLEU} & \textbf{R-BLEU} & \textbf{PBSMT-BLEU} & \textbf{E-4-gram} & \textbf{BLEU} \\ 
\midrule
Assamese (as) & 50000 & 1.44 & 8.19 & 6.03 & 484.92 & 66.36 \\
Mizo (lus) & 50000 & 2.08 & 15.77 & 11.10 & 1141.87 & 33.30 \\
Manipuri (mni) & 21687 & 1.24 & 19.91 & 14.85 & 330.78 & 69.75  \\
Khasi (kha) & 24000 & 2.00 & 4.43 & 7.67 & 727.82 & 20.72\\
\midrule
\textbf{Lang} &  & \textbf{D-chrF} & \textbf{R-chrF} & \textbf{PBSMT-chrF} &  &  \\ 
\midrule
Assamese (as) &  & 11.59 & 24.89 & 26.30 &  & 75.88 \\
Mizo (lus) &  & 14.87 & 31.59 & 31.11 &  & 52.74 \\
Manipuri (mni) &  & 13.07 & 38.84 & 42.69 &  & 78.16 \\
Khasi (kha) &  & 19.61 & 19.51 & 31.82 &  & 43.34 \\
\midrule
\textbf{Lang} &  & \textbf{D-chrF++} & \textbf{R-chrF++} & \textbf{PBSMT-chrF++} &  &  \\ 
\midrule
Assamese (as) &  & 9.47 & 22.86 & 24.39 &  &  \\
Mizo (lus) &  & 13.13 & 30.46 & 29.57 &  &  \\
Manipuri (mni) &  & 10.50 & 36.90 & 40.46 &  &   \\
Khasi (kha) &  & 17.87 & 18.05 & 30.31 &  & \\
\bottomrule
\end{tabular}
}
\caption{Translation Performance Metrics for Indic language pairs. Exposure scores are calculated by counting on the high-resource side of the language pairs. BLEU and chrF scores are taken from the best performance in WMT 2023 shared task\cite{pal-etal-2023-findings}. \label{tb:indic}}
\label{tab:indic-languages}
\end{table}

\section{Implementation Details of Metrics\label{ch:metric_imp_detail}}
\subsection{Implementation details on Pretraining Exposure (E)}

In our implementation, we utilize the \textbf{infini-gram} engine \cite{liu2025infinigramscalingunboundedngram} to index the pretraining dataset for fast 4-gram count retrieval. The implementation details are as follows:
\begin{enumerate}
    \item \textbf{Dataset retrieval: } We retrieved a subset of public bitext which is a part of the pretraining dataset of NLLB from the official NLLB GitHub repository\footnote{\url{https://github.com/facebookresearch/fairseq/blob/nllb/examples/nllb/data/download_parallel_corpora.py}}.
    \item \textbf{Indexing: } We utilized infini-gram engine to index all the bitext retrieved, using mBART-50 tokenizer \cite{mbart} for gram separation.
    \item \textbf{Counting: } For a target test corpus, we tokenized all sentences in target language in the test corpus using the same mBART-50 tokenizer, splited into all possible 4-grams, and used indexed infini-gram engine to retrieve the count of each 4-gram in the pretraining dataset.
    \item \textbf{Report: } Take the mean value for all possible 4-grams and report as result.
\end{enumerate}

\subsection{Tokenization policy on PBSMT training and BLEU calculation}

To adapt variability in nature of various languages, we define the following policies in our calculation of BLEU scores:

\begin{itemize}
    \item All BLEU scores are \textbf{average sentence BLEU scores} computed using the \verb|sentence_bleu| implementation from SacreBLEU \cite{post2018clarityreportingbleuscores}. To ensure consistent evaluation, we use the internal default setting of the \verb|sentence_bleu| method with exponential smoothing except changing the tokenizer for calculating BLEU depends on the nature of different languages.
    \item For most space-separated languages, we utilize the built-in default \verb|13a| tokenizer adapted by sacreBLEU, which is also the WMT standard tokenizer and suitable for space-separated languages.
    \item For Chinese languages in both high-resource languages group and Formosan Mandarin group, we utilize the built-in \verb|zh| tokenizer adapted by sacreBLEU, which does character-wise separation on Chinese characters but preserve word structures on other space-separated languages.
    \item For Japanese in high-resource languages, we use built-in \verb|ja-mecab| tokenizer adapted by sacreBLEU.
    \item For Akkadian ancient language, we utilize the built-in \verb|char| tokenizer adapted by sacreBLEU which does character-wise separation.
\end{itemize}

We also defined the tokenization policy on training phrase-based translation systems:

\begin{itemize}
    \item We used a Docker-powered open-source project called \verb|Moseskit| \cite{moseskit} for running Moses PBSMT \cite{lample2018phrasebasedneuralunsupervised}. We keep the tunable configurations default to ensure consistent evaluation, including using 5-gram language model and default tokenizer that does space-separation. To maintain evaluation consistency, we use \verb|13a| tokenizer in SacreBLEU to calculate average sentence BLEU between PBSMT prediction and test dataset ground truths.
    \item For non-space-separating languages like Chinese, we use MBart tokenizer \cite{mbart} to separate words for PBSMT training. We have conducted experiments to test the PBSMT performances on whether we use MBart in non-space-separating language side only, or both source and target languages in Formosan language corpus, table \ref{tab:formosan-mbart-tokenize-pbsmt-performance} shows that PBSMT generally performs better if apply MBart in one side only. We use \verb|zh| tokenizer for Chinese, \verb|ja-mecab| tokenizer for Japanese, and \verb|char| tokenizer for Akkadian in sentence BLEU evaluation for consistency. 
    \item The default argument \verb|kndiscount| for applying Kneser-Ney Discounting on PBSMT may fail when either the dataset length or the lexical diversity (number of unique characters or words) being too small. In this case, we switch to fallback \verb|wbdiscount| to use Witten-Bell Discounting, and reduce the tuning set size to 50 if the training dataset is small as only a few hundred samples.
\end{itemize}

\begin{table}[h]
    \centering
    \resizebox{0.7\linewidth}{!}{
        \begin{tabular}{@{}lcc@{}}
        \toprule
           \multirow{2}{*}{\textbf{Lang}}  & \multirow{2}{*}{\textbf{$\frac{\text{zh only}}{\text{BLEU / chrF}}$}} & \multirow{2}{*}{\textbf{$\frac{\text{Both side}}{\text{BLEU / chrF}}$}} \\\\
        \midrule
Sakizaya (ais) & 9.17 / 8.17 & 6.78 / 7.43 \\
Amis (ami) & 8.97 / 6.38 & 6.73 / 5.96 \\
Bunun (bnn) & 13.97 / 10.60 & 12.36 / 11.69  \\
Kavalan (ckv) & 19.81 / 16.03 & 16.26 / 14.79 \\
Rukai (dru) & 11.39 / 7.79 & 10.11 / 9.08 \\
Paiwan (pwn) & 9.77 / 9.82 & 9.02 / 9.82 \\
Puyuma (pyu) & 11.22 / 10.15 & 8.99 / 9.35 \\
Seediq (sdq) & 8.03 / 8.61 & 6.62 / 8.20 \\
Thao (ssf) & 15.70 / 12.87 & 14.75 / 13.78 \\
Saaroa (sxr) & 8.41 / 6.97 & 8.08 / 9.21 \\
Yami (tao) & 11.22 / 9.19 & 9.14 / 8.82 \\
Atayal (tay) & 8.96 / 6.39 & 6.69 / 5.92 \\
Truku (trv) & 9.74 / 7.95 & 8.73 / 8.47 \\
Tsou (tsu) & 9.57 / 9.27 & 7.78 / 9.63 \\
Kanakanavu (xnb) & 16.57 / 13.92 & 14.14 / 14.62 \\
Saisiyat (xsy) & 11.72 / 10.40 & 10.45 / 10.64 \\
        \bottomrule
        \end{tabular}
    } 
    \caption{PBSMT Scores for Formosan language pairs: translate to high-resource (zh) direction. The two columns shows the performance difference of using MBart as tokenizer on Chinese side only and using MBart on both sides.}
    \label{tab:formosan-mbart-tokenize-pbsmt-performance}
\end{table}

\section{$R$-score and SMT Correlation\label{ch:smt_correlation}}

\begin{figure}[h]
    \centering
    \includegraphics[width=0.95\linewidth]{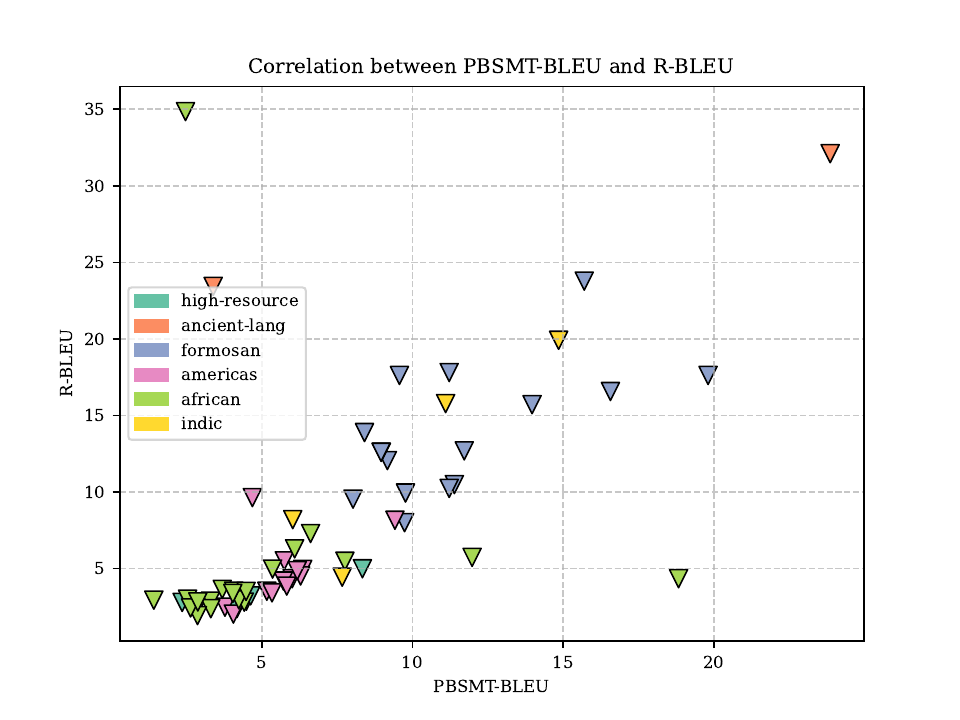}
    \caption{Scores of PBSMT versus R-BLEU by language groups.}
    \label{pbsmt_bleu}
\end{figure}

We plot a correlation map between PBSMT \cite{koehn2003statistical, moseskit} and $R$-score (Figure~\ref{pbsmt_bleu}). $R$-score provides a computationally efficient, parameter-free proxy for PBSMT-like signals and appears to align reasonably well with PBSMT in practice. In Figure~\ref{pbsmt_bleu}, PBSMT and R-BLEU are positively correlated (Pearson $r = 0.592$), and excluding three outliers from the African and Ancient language group (Zulu \texttt{zul}, Nigerian-Pidgin \texttt{pcm}, and Egyptian \texttt{egy}) increases the coefficient to $0.902$. 

A manual inspection further suggests that \texttt{zul} and \texttt{egy} contain many train/test pairs with identical target-side sentences but mildly different source-side sentences; this data characteristic may inflate retrieval-based scores ($R$-score often near 100) while making phrase extraction in PBSMT less stable.

\section{Computational Cost\label{ch:computecost}}

\textbf{D and R Scores: } These metrics run in comparable execution time since they all iterate every possible train-test pairs. For a dataset with approximately 1000 test samples and 15000 training samples, where each sample has an average of 25 tokens, the process runs for roughly 2 minutes with 32 threads, which is 0.9 CPU-hours on two AMD EPYC 7282 16-Core Processors.\\

\textbf{PBSMT: } In our experiment, the Moses PBSMT training process was configured to run on 24 CPU threads on an AMD Ryzen Threadripper 3960X 24-Core Processor. A training corpus with around 5000 parallel sentences with around 100k total tokens runs in approximately 2 hours.\\

\textbf{$E$-score: } Utilizing the infinigram engine \cite{liu2025infinigramscalingunboundedngram}, the indexing process on an approximately 30GB bitext pretraining dataset costs roughly around 3 hours. The indexing process is configured to use 16 threads of two AMD EPYC 7282 16-Core Processors CPU, 32GB of memory, and 524288 open-file limit, with the custom MBart tokenizer with unsigned 32-bit integer token datatype. The final index folder occupies around 60GB of storage space. Calculating $E$-score for a 3000 lines test dataset with roughly 15000 unique 4 grams cost around 30 seconds.

\end{document}

%% file: macro.tex
\usepackage{enumitem}
\usepackage{multirow}
\usepackage{xcolor}
\usepackage{xspace}
\usepackage{amsmath}

\usepackage[capitalize,noabbrev]{cleveref}
\crefname{chapter}{Chapter}{Chapters}
\crefname{section}{\S}{\S\S}
\Crefname{section}{\S}{\S\S}
\crefname{table}{Table}{Tables}
\crefname{figure}{Figure}{Figures}
\crefname{algorithm}{Algorithm}{}
\crefname{equation}{Eq.}{}
\crefname{appendix}{Appendix}{}
\crefformat{section}{\S#2#1#3}
\usepackage{float}

\usepackage{booktabs}

\newif\ifdraft
\drafttrue

\usepackage{todonotes}

\definecolor{tticblue}{RGB}{0, 94, 184}